\documentclass[sigconf]{acmart}
\AtBeginDocument{%
  \providecommand\BibTeX{{%
    \normalfont B\kern-0.5em{\scshape i\kern-0.25em b}\kern-0.8em\TeX}}}

\copyrightyear{2023}
\acmYear{2023}
\setcopyright{rightsretained}
\acmConference[CHI '23]{Proceedings of the 2023 CHI Conference on Human Factors in Computing Systems}{April 23--28, 2023}{Hamburg, Germany}
\acmBooktitle{Proceedings of the 2023 CHI Conference on Human Factors in Computing Systems (CHI '23), April 23--28, 2023, Hamburg, Germany}\acmDOI{10.1145/3544548.3581351}
\acmISBN{978-1-4503-9421-5/23/04}

\acmConference[Conference CHI '23]{Make sure to enter the correct
  conference title from your rights confirmation emai}{April 23-28,
  2023}{Hamburg, Germany}
\usepackage{xspace}

\newcommand{\xhdr}[1]{{\noindent\bfseries #1.}}
\newcommand{\xhdrq}[1]{{\noindent\bfseries #1?}}

\newcommand{\dispatch}{\textsc{dispatch}\xspace}
\newcommand{\msg}{message\xspace}

\definecolor{turquoise}{RGB}{69, 181, 170}
\definecolor{azure}{rgb}{0.0, 0.5, 1.0}

\newif\ifshowcomments
\showcommentsfalse

\newcommand\todo{\textcolor{red}}
\newcommand\cut[1]{}
\newcommand{\bn}[1]{{\textcolor{turquoise}{[Ben: #1]}}}  
\newcommand{\ly}[1]{{\textcolor{azure}{[Liye: #1]}}} 
 
\newcommand{\sk}[1]{{\textcolor{orange}{#1}}}
\newcommand{\rev}[1]{{\textcolor{blue}{#1}}} 
\ifshowcomments
\else
\renewcommand\todo[1]{}
\renewcommand{\bn}[1]{}
\renewcommand{\ly}[1]{}
\renewcommand{\sk}[1]{}
\renewcommand{\rev}[1]{\textcolor{black}{#1}}
\fi

\usepackage{subfig} 

\begin{document}

\title[Dispatch]{Comparing Sentence-Level Suggestions to Message-Level Suggestions in AI-Mediated Communication}



\author{Liye Fu}
\authornote{Work done while at Cornell University.}
\email{liye.fu@thomsonreuters.com}
\orcid{0000-0001-7989-6839}
\affiliation{%
  \institution{Thomson Reuters Labs}
  \city{Toronto}
  \country{Canada}
}

\author{Benjamin Newman}
\email{benjaminn@allenai.org}
\orcid{0000-0003-3552-8676}
\affiliation{%
  \institution{Allen Institute for Artificial Intelligence}
  \city{Seattle}
  \country{USA}
}

\author{Maurice Jakesch}
\email{mpj32@cornell.edu}
\orcid{0000-0002-2642-3322}
\affiliation{%
  \institution{Cornell University}
  \city{Ithaca}
  \country{USA}
}

\author{Sarah Kreps}
\email{sarah.kreps@cornell.edu}
\orcid{0000-0002-0924-4234}
\affiliation{%
  \institution{Cornell University}
  \city{Ithaca}
  \country{USA}
}



\begin{abstract}

Traditionally, writing assistance systems have focused on short or even single-word suggestions. Recently, large language models like GPT-3 have made it possible to generate significantly longer natural-sounding suggestions, offering more advanced assistance opportunities.
This study explores the trade-offs between sentence- vs. message-level suggestions for AI-mediated communication. We recruited 120 participants to act as staffers from legislators' offices who often need to respond to large volumes of constituent concerns. Participants were asked to reply to emails with different types of assistance. The results show that participants receiving message-level suggestions responded faster and were more satisfied with the experience, as they mainly edited the suggested drafts. In addition, the texts they wrote were evaluated as more helpful by others. In comparison, participants receiving sentence-level assistance retained a higher sense of agency, but took longer for the task as they needed to plan the flow of their responses and decide when to use suggestions. Our findings have implications for designing task-appropriate communication assistance systems. 

\end{abstract}

\maketitle

\section{Introduction}
\label{sec:intro}

Traditional communication assistance systems have generally focused on short suggestions to improve input efficiency. 
With the emergence of large language models such as GPT-3 \cite{brown2020language}, it has become possible to generate significantly longer natural-sounding text suggestions, opening up opportunities to design more advanced writing assistance to help humans with more complex tasks in more substantial ways \rev{\cite{Lee2022coauthor,wodzak_can_2022}}.
Such assistance can be especially helpful in communication scenarios in which a single point of contact needs to manage large volumes of correspondence, e.g., customer service representatives addressing customers’ queries, professors attending to students’ emails, as well as elected officials responding to their constituents’ concerns.

The generative capabilities of current models enable a wide range of possibilities for designing assistance systems. In this work, we explore two writing assistant design choices, namely sentence-level and \msg-level text suggestions, and empirically analyze the trade-offs between them in email communication.\footnote{In our context, \msg-level suggestion means a full draft for responding to an email.} We consider the practical scenario of staffers from legislators' offices responding to vast amounts of constituents' concerns as the context for our study. In this context, the volume of correspondence can become overwhelming, making intelligent assistance especially needed \cite{the_opengov_foundation_voicemails_2017,congressional_management_foundation_communicating_2022}. 
At the same time, the high-stakes nature of political communication calls for more careful and comprehensive research to better understand the potential benefits and risks of any type of technical assistance that may be introduced to the existing workflow. 
%


To advance our understanding of different assistance options, we develop \dispatch (Section~\ref{sec:dispatch}), an application that can serve as a platform to simulate the process of a staffer responding to constituents’ emails, 
allowing us to design and set up an online experiment to test different types of suggestions (Section~\ref{sec:evaluation}).
We recruited 120 participants to act as staffers from legislators' offices to respond to three emails expressing constituents' concerns under three different conditions: 40 participants received no assistance, 40 received sentence-level suggestions, and 40 \msg-level suggestions, with both types of suggestions generated by GPT-3. 

By observing participants' interactions with text suggestions and surveying their perceptions of the assistance they received, we are able to compare how sentence-level suggestions and \msg-level suggestions may affect the participants' writing experience as well as the eventual responses produced. 
The results show that participants who receive \msg-level suggestions generally found the suggested drafts natural and mainly edited on top of them. 
They were able to finish their responses significantly faster, demonstrating an increased level of efficiency in responding, and were generally more satisfied with the assistance they received. 
In comparison, participants receiving sentence-level assistance took longer since they still had to plan for the key points to cover in their responses, while deciding where to trigger and use suggestions. 
While they retained a higher sense of agency, they spent longer time and reported lower levels of satisfaction, demonstrating the challenging nature of designing assistance systems with finer-grained control. 
%
%
This discrepancy points to the need to take receivers' perspectives into account when designing and introducing assistance systems into the workflow in communication circumstance where trust is highly valued. 
We discuss the implications of our experimental results for designing task-appropriate communication assistance systems (Section~\ref{sec:implications}).

\section{Related Work}
\label{sec:related}

\begin{figure*}[!h]
    \centering
    \includegraphics[width=0.8\textwidth]{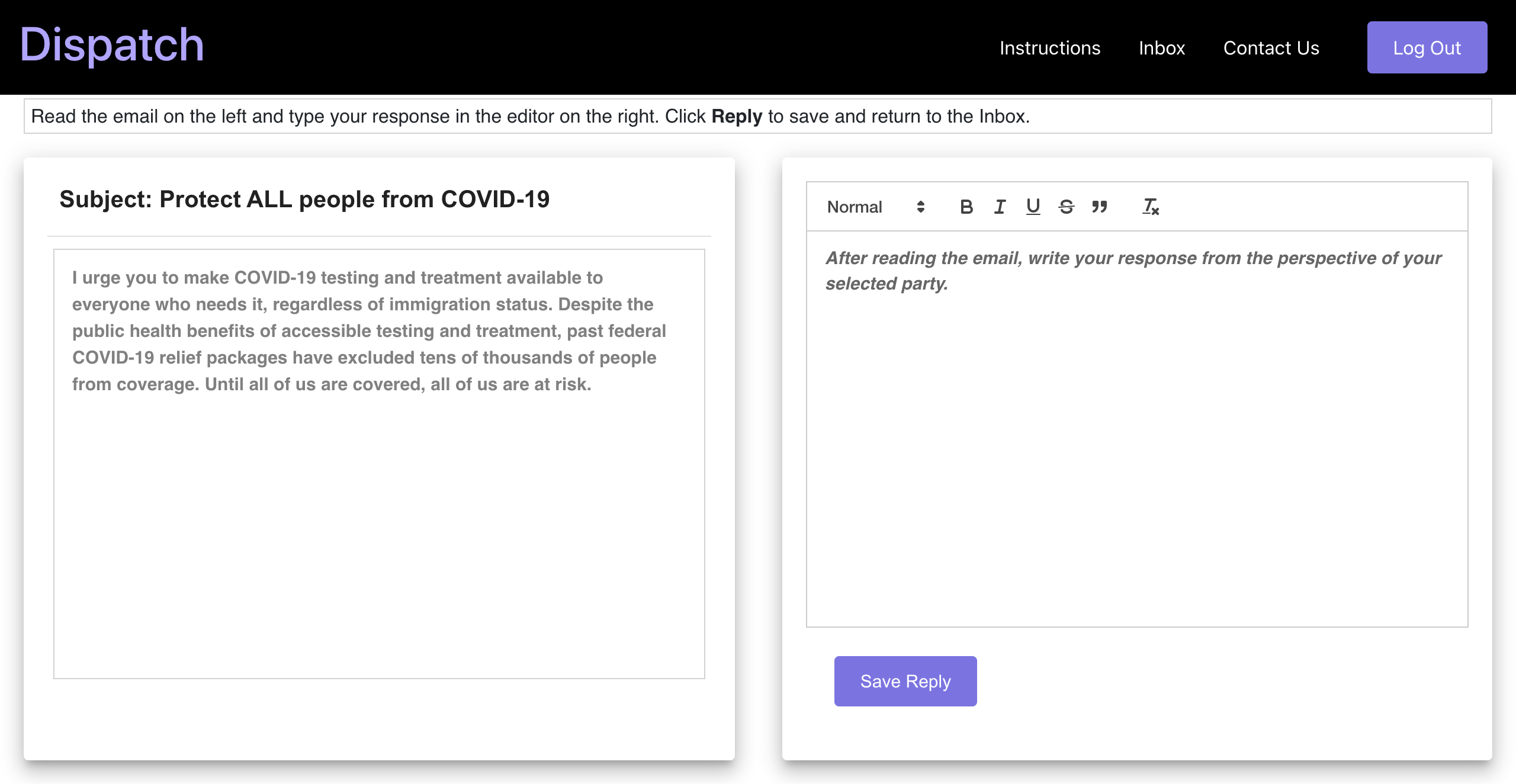}
    \caption{The basic \dispatch interface. The subject and body of the constituent letter are displayed on the left, while an editor for drafting the response is provided on the right.}
    \label{fig:interface-control}
\end{figure*}

\subsection{Advances in AI text generation}
Advances in machine learning have led to a new generation of language models \cite{bommasani2021opportunities} capable of producing text indistinguishable from human-written content \cite{jakesch2022human, kreps2022all}. Enabled by improvements in computer hardware and the transformer neural network architecture \cite{vaswani2017attention}, models like GPT-3 \cite{brown2020language} have attracted attention for their ability to generate text that mimics the style and substance of the inputs. Cautious voices have warned about the ethical and social risks of harm from large language models \cite{weidinger2021ethical, weidinger2022taxonomy}, ranging from discrimination and exclusion \cite{huang2019reducing, brown2020language, nozza2021honest} to misinformation \cite{kreps2022all, lin2021truthfulqa, rae2021scaling} and environmental \cite{strubell2019energy} and socioeconomic harms \cite{bender2021dangers}.  

However, these same technologies have potential to usher in a range of beneficial real-world applications \cite{bommasani2021opportunities}. These models have the potential to aid in journalism, curate weather and financial reports, and write customer-service responses, with particular value in domains where the task is either repetitive or has high volume writing requirements.   

Building on the core technological foundation, more recent research in computer science, HCI, and linguistics has focused on input efficiency, often by exploiting linguistic information to speed up the writing process \cite{Kristensson2014}.
Early predictive text systems such as T9 relied on word frequencies to suggest word continuations~\cite{James2000}.
More advanced systems combine behavioral data \cite{Goodman2002} with information at the sentence level \cite{Vertanen2015velocitap} to predict users' intentions to complete entire phrases or sentences~\cite{Arnold2016phrases_vs_words, Buschek2021emails}. To increase the likelihood of a matching suggestion, systems like today's smartphone keyboards provide multiple suggestions in parallel \cite{Kannan2016smartreply}. Some systems, like Google's \textit{Smart Compose}~\cite{Chen2019smartcompose} use the estimated utility or probability of acceptance to determine whether suggestions should be shown. 

Writing assistants usually provide short, or single-word suggestions only \cite{Dunlop2012, Fowler2015, Quinn2016chi}, with the assumption that for longer suggestions, the time required to evaluate the suggestion may distract from or even slow down the process of composition. Indeed, prior studies have suggested that writing suggestions can reduce typing performance and deteriorate the user experience \cite{Banovic2019mobilehci, Buschek2018researchime, Palin2019}. However, this may change with advances in the quality of text generated by language models \cite{bommasani2021opportunities, brown2020language}. 
Massive transformer neural networks \cite{vaswani2017attention} that manage to capture more complex user intents may be able to provide higher quality suggestions that are likely to be useful, thus reducing the relative cost of evaluation. 
In addition, these models may provide ideas and inspiration \cite{Lee2022coauthor, Singh2022elephant, Yuan2022wordcraft} beyond simply increasing text input efficiency. 

\subsection{The use of technology in political communication}

Democratic accountability implies communication between elected leaders and constituents, wherein constituents write to express their concerns and preferences and elected leaders respond to articulate how they plan to or have addressed these preferences \citep{hertel-fernandez_mildenberger_stokes_2019, grose_explanations_2015}. 
As technology has made it easier to contact members of Congress, for example through representative websites with ``contact'' buttons and civil society organizations providing email templates, the volume of mail has increased considerably, making the task of meaningfully processing and responding to correspondence more difficult \citep{cmf-volume-2017}. 

\rev{Social media has provided one way for elected leaders to correspond with large numbers of constituents to help understand their concerns and explain leaders' positions \citep{barbera2019leads}. While voters have a number of tools available to contact their legislators and craft letters and emails, legislators lack analogous tools to help craft responses. This contributes to legislative staffers, tasked with the responsibility of reading and responding to the large volume of incoming mail, being less responsive to communications from some groups \citep{barbera2019leads}.} 

It is in this space where AI-mediated communication could potentially be fruitful.
Research on the use of language models or writing assistants within the political process has thus far been limited; however, we can look to work on the political ramifications of social media feeds and recommender systems \cite{zhuravskaya2020political} to offer clues about the possible impact of these advancements.
Despite initial excitement about these technologies' democratic potential \cite{khondker2011role}, scholars have identified the potential for these technologies to become the subject of powerful political and commercial interests \cite{bradshaw2017troops} that may undermine democratic institutions \cite{aral2019protecting}.
Even unintentionally, design choices related to algorithmic optimization may lead to self-reinforcing opinion dynamics \cite{bruns2019filter}.
Similarly, language model writing assistants also have to be designed carefully if they are to be used for for political communication.
Prior work has shown that when language models perform poorly (e.g., produce repetitive outputs), they may corrode  constituents' trust in their elected representatives \cite{kreps_ai-mediated_2022}.
This suggests that humans curating model outputs, as well as continued improvements in providing diverse types of high quality suggestions---from short, single-word suggestions to paragraph or even longer length---could alleviate these impediments.

Building on this research, it appears that human-crafted responses, facilitated by language models that offer suggestions, could play a role in facilitating thoughtful interaction. \citet{kreps2022all} have shown that people are largely unable to detect political news generated by recent generations of language models, suggesting that these models, assisted by a human in the loop, would be effective at generating content that bridges elected leaders with their constituents. Nonetheless, any use of language models for political purposes will need to be carefully assessed in the light of their political consequences that research such as this uncovers.


\section{Designing Dispatch}
\label{sec:dispatch}

To understand the trade-offs between \msg-level and sentence-level suggestions, we built \dispatch, a platform to simulate the scenario of staffers from legislative offices responding to constituents' concerns.
The basic \dispatch interface is shown in Figure~\ref{fig:interface-control}.
A letter from a constituent is displayed on the left, while an editor for drafting the response is provided on the right. The editor supports all typical actions for writing, e.g., typing, deleting, and cursor movements. 

On top of this basic interface, we build two different versions of \dispatch, one that offers sentence-level suggestions and another that offers \msg-level suggestions.

\begin{figure*}[!h]
    \centering
    \includegraphics[width=0.8\textwidth]{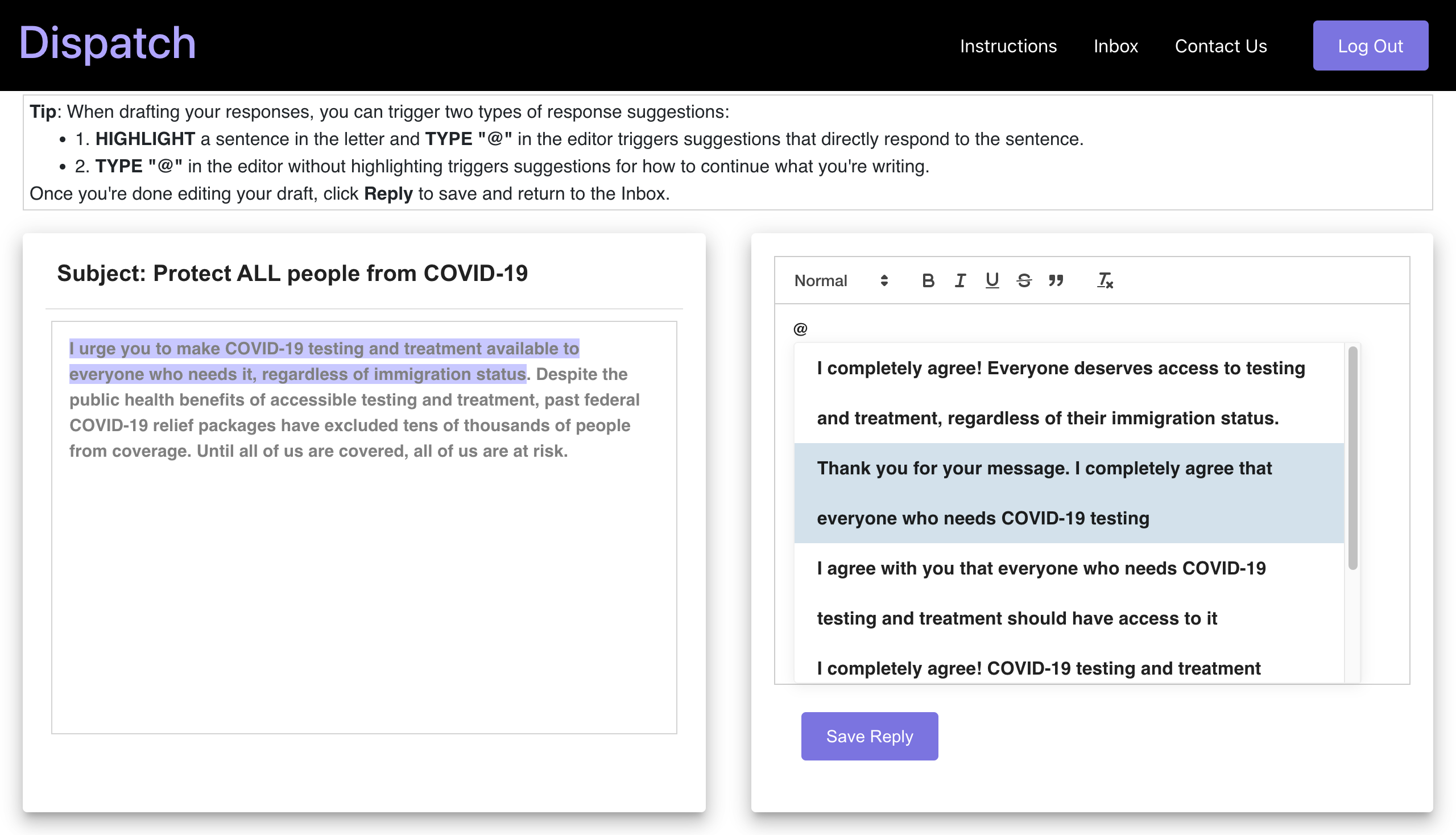}
    \caption{Users can trigger sentence-level suggestions by typing ``@" in the editor. Five candidate suggestions will be shown in a drop-down menu. If the user has highlighted a sentence in the email (as shown in the figure), response suggestions are provided. Otherwise, suggestions are given to continue the current draft.}
    \label{fig:interface-sentence}
\end{figure*}

\xhdr{Sentence-level suggestions} We allow users to trigger two types of sentence-level suggestions. 
First, users can receive suggestions responding to specific points raised in the constituent letter by highlighting the sentence to respond to and then typing ``@" in the editor (Figure~\ref{fig:interface-sentence}).
Second, users can trigger suggestions to continue the text they have already written\footnote{We use 30 tokens before the current cursor position as the prompt.} by typing ``@" without  highlighting any text. 
In both cases, users are presented with a drop-down menu displaying five candidate suggestions, from which they \rev{can} select one \rev{or none}.

\begin{figure*}[!h]
    \centering
    \includegraphics[width=0.8\textwidth]{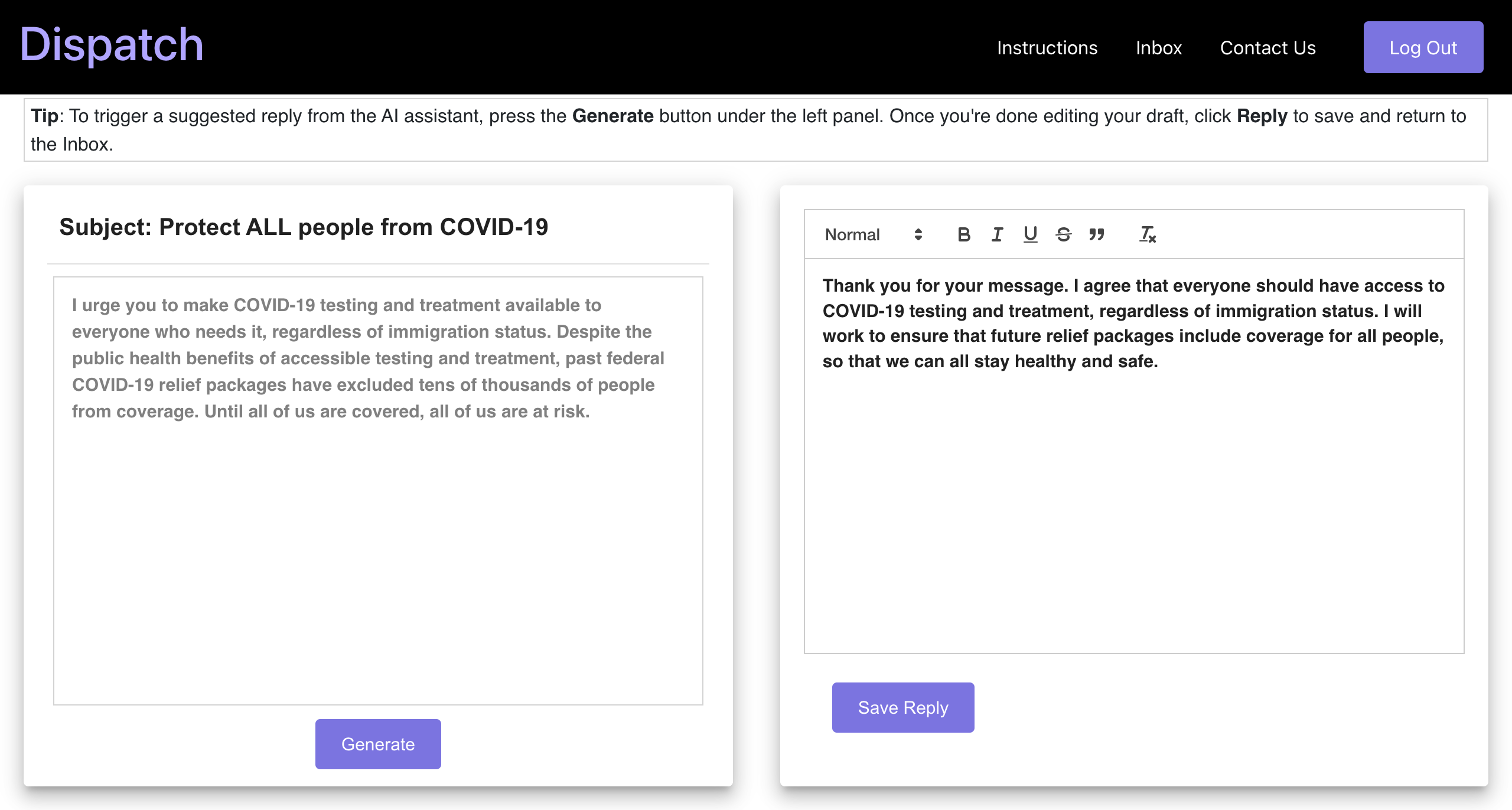}
    \caption{Users can trigger a message-level suggestion by clicking the ``Generate" button. The response suggestion will be directly loaded in the editor on the left as shown.}
    \label{fig:interface-email}
\end{figure*}

\xhdr{Message-level suggestions} In the \msg-level suggestion interface, users can click on the ``Generate" button to obtain a full response draft. The suggestion is directly loaded into the editor (Figure~\ref{fig:interface-email}) for users to make further edit.\footnote{\rev{Note that the two types of suggestions are generated independently, i.e., the sentence-level suggestions are not intentionally a subset of the message-level suggestions although there can be coincidental overlaps depending on how participants trigger suggestions.}}

\section{System Evaluation}
\label{sec:evaluation}

\subsection{Experiment setup} 
\label{sec:experiment}

\xhdr{Constituents' letters} As proxies for constituents' emails, we sample open letters delivered to elected officials in the United States through Resistbot, a service that advertises the ability to compose and send letters to legislators in less than two minutes.\footnote{\url{https://resist.bot}.} We obtain images of these open letters by retrieving tweets published by @openletterbot\footnote{\url{https://twitter.com/openletterbot}.} using the Twitter API and extract the contents of the letters using Python Tesseract.\footnote{\url{https://github.com/madmaze/pytesseract}.} For each letter, we keep only the content of the letter, removing the sender's first name and the state they are from, if applicable. In addition, we only consider letters that are sent by multiple people to ensure that the letters are \rev{representative but} not personally identifiable.

To select letters to use as prompts for participants to respond, we consider topics that are generally relatable but not overtly polarizing. 
\rev{Our choice is based on two considerations. First, common concerns constitute a significant portion of the emails legislators need to address, as we observe a substantial number of near-duplicate open letters expressing similar concerns. Second, more general topics would make the task manageable to our participants who might not be familiar with very niche topics. 
While staffers would also have to respond to more specific questions, focusing on common ones is sufficient for exploring the difference between the two types of suggestion types.}
The twelve letters we select span topics such as health insurance, climate change, and COVID relief policies.\footnote{The full set of letters is included in the Supplementary Materials.} The average length of the selected letters is 97 words.

\begin{figure*}[!h]
    \centering
    \includegraphics[width=0.9\linewidth]{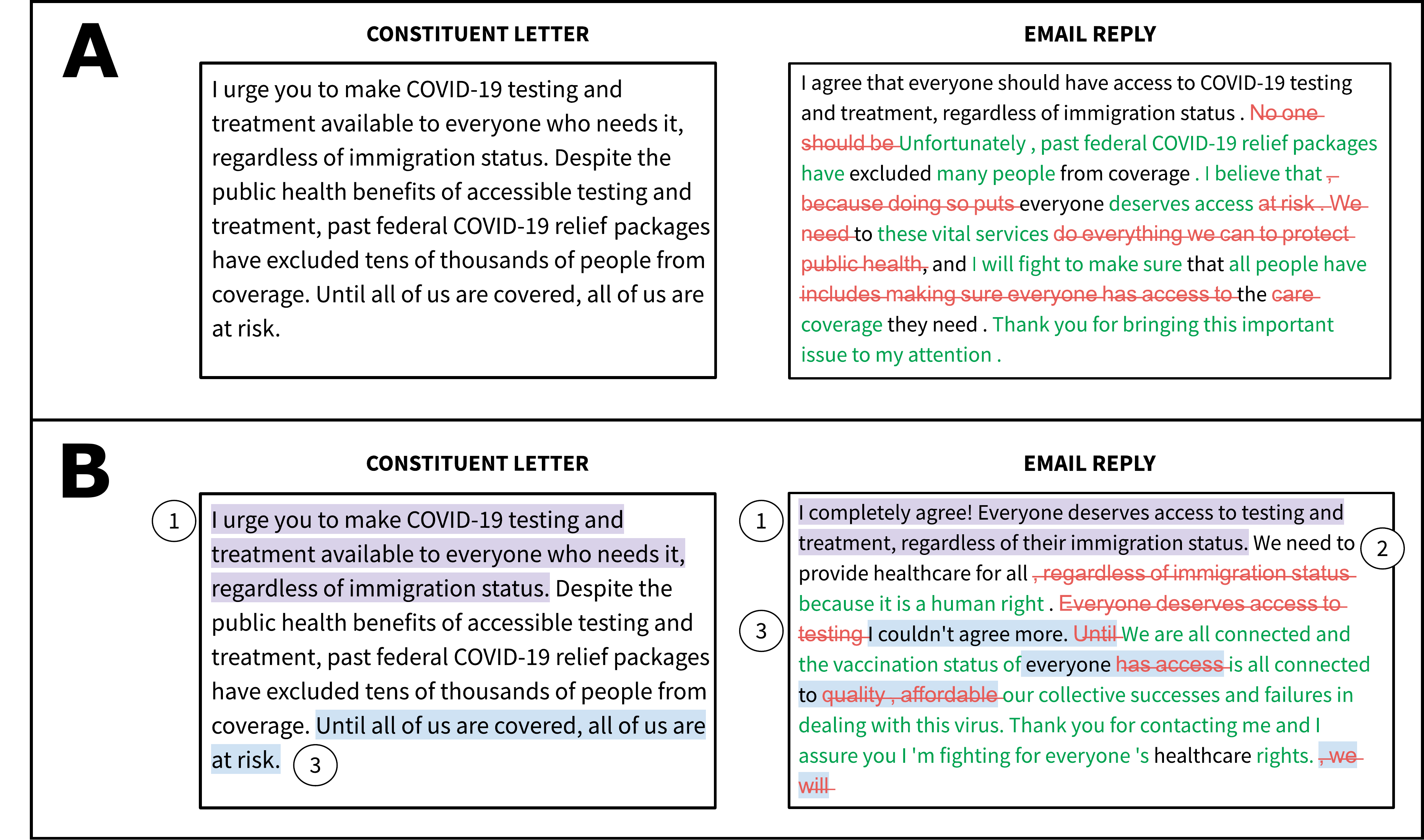}
    \caption{\textbf{A.} Sample constituent letter and reply from the \textbf{message-level suggestions} condition. Black text was suggested, red struck-out text was removed from the suggestion, and green text was added by the participant. 
    \textbf{B.} Sample reply and constituent letter from the \textbf{sentence-level suggestions} condition.  Each circled number represents a place where the participant prompted a suggestion. (1) and (3) were prompted with letter text highlighted. The highlighted text and the suggestion given were highlighted in purple in (1) and light blue in (3). Suggestion (2) was a continuation suggestion, prompted without any text in the constituent letter highlighted. Best viewed in color.}
    \label{fig:diffs}
\end{figure*}

\xhdr{Experimental conditions} In our experiment, we randomly assign participants to one of three conditions:
\begin{enumerate}
    \item {\bf Control}: participants respond to emails in their own writing, with no response suggestions.
    \item {\bf Sentence-level suggestions}: participants have the option to trigger sentence-level suggestions, with the option to accept and edit them. 
    \item {\bf Message-level suggestions}: participants have the option to trigger an automatically generated full email response draft to edit.
\end{enumerate}

\xhdr{Generation model} We use GPT-3 (specifically, the \textsc{text-davinci-002} model without any fine-tuning) to generate suggestions.
We set {\it max\_tokens} to 200 when generating full email drafts and 20 when generating sentence-level suggestions. Under both conditions, we set {\it temperature} to 0.7 and {\it top\_p} to 0.96 throughout.
Our application was reviewed and approved by OpenAI before launching the experiment. 

\xhdr{Participants} We recruit 40 participants for each experimental condition via the platform Prolific \cite{palan2018prolific}.
We only consider participants who are located in the United States, fluent in English, and have listed ``politics'' as one of their hobbies.
Each eligible participant is allowed to take part in at most one condition. 
We pay \$5.00 for each task session based on an estimated completion time of 20 minutes. The actual completion time in each condition is shown in Figure~\ref{fig:completion-time}. 
The experiment received Institutional Review Board approval from Cornell University.

\subsection{Writers' experience}
\label{sec:experience}

To understand the participants' writing processes, we track both the overall time they spend on the task, the suggestions they trigger, and the responses they submit. 
Figure~\ref{fig:diffs} shows sample responses annotated with participants' interactions with the text suggestions. 

\begin{figure}[!t]
    \centering
    \includegraphics[width=0.45\textwidth]{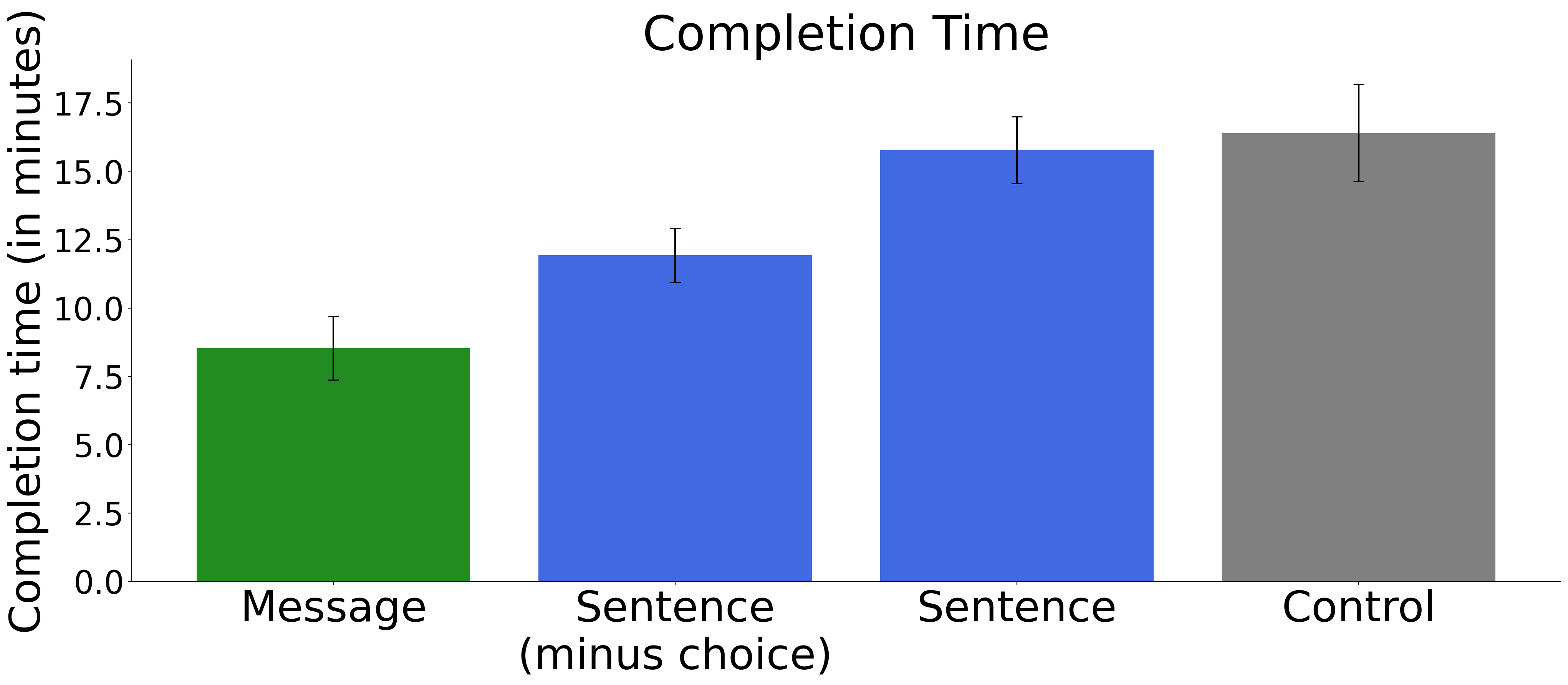}
    \caption{Participants in the message-level suggestions condition tend to finish the task significantly faster than both participants in the control condition and participants who can trigger sentence-level suggestions.}
    \label{fig:completion-time}
\end{figure}

\xhdr{Completion time} We compare the average completion time across three experimental conditions to explore whether offering writing assistance helps participants respond to emails more efficiently (Figure~\ref{fig:completion-time}). 
We find that participants in the message-level suggestions took the least time to complete the task. On average, they finished responding to all three emails in a mean time of 8.53 minutes, which is significantly faster than both participants in the control group ($M=16.40$, $t_{(78)}=-3.71$, $p<0.001$), and participants in the sentence-level suggestions condition ($M=15.77$, $t_{(78)}=-4.30$, $p<0.001$).\footnote{\rev{Throughout, we use independent-samples t-test with Bonferroni correction. In this subsection, as we make four comparisons, a Bonferroni corrected alpha level of $0.0125$ is used.}}
This suggests that offering drafting suggestions has the potential to help people write responses faster.\footnote{We recognize that the faster response time may be partially attributed to shorter replies. However, the trend remains similar even if response lengths are taken into account, i.e., if we consider time taken per word.}
However, we do not observe a significant difference between the completion times of the sentence-level suggestions condition and the control condition, potentially because the time saved from generating ideas and typing sentences was offset by time spent in choosing between suggestions as well as time wasted when generated suggestions were not good enough to be used.  
When the time taken to select suggestions is removed, the total writing time is closer to the message-level suggestions condition ($M=11.93$), though the difference between sentence-level and control is still not statistically significant at the Bonferroni corrected alpha level of $0.0125$, with $t_{(78)}=2.21$, $p=0.030$.

\xhdr{Interactions with the suggestions} A central question across the conditions is how participants used the generated suggestions. 
We consider each response writing process---starting when the participant views a letter in the interface and ending when they save their reply---as one interaction. 
Most participants have three interactions, one for each letter, and when there is more than one, we choose the interaction that resulted in the saved reply.
This gives 120 recorded interactions (i.e., 40 participants $\times$ 3 interactions) per experimental condition.

\begin{figure}
    \centering
    \includegraphics[width=0.4\textwidth]{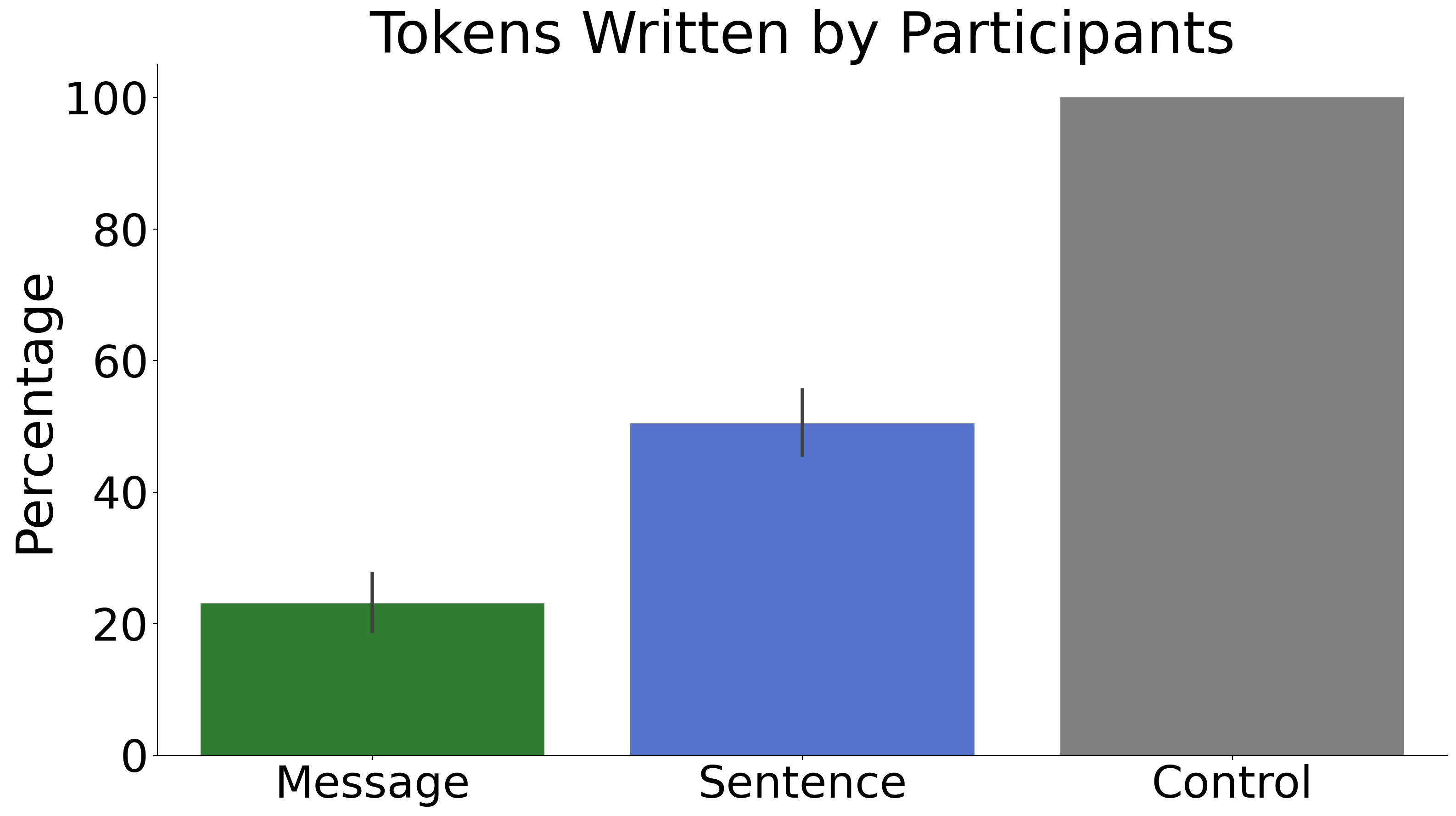}
    \caption{Average percentage of participant-written text across all the conditions. All of the control replies are participant-written while almost a quarter of the message-level replies and half of all sentence-level replies are participant-written.}
    \label{fig:eval-pct-human-all}
\end{figure}

In the \msg-level suggestions condition, every participant queried the model for at least one suggestion.\footnote{In 11 interactions, a participant queried twice, and more (three, five, and six suggestions) were queried only once each.}
%
Once a participant received a suggestion, they often stuck closely to it: on average, 75.75\% of the tokens in the final replies came from the suggestions, while the other 24.25\% were added by the participants (Figure~\ref{fig:eval-pct-human-all}).
Furthermore, in 31 (25.8\%) of the interactions, participants accepted the suggestions without editing them, and only in 2 (1.67\%) did a participant choose to completely remove a suggestion and write their own response.

The sentence-level suggestions condition has more complex interaction patterns because participants were expected to query for suggestions multiple times and in two distinct ways (either with or without highlighting text from the letter to respond to).
Because of this, participants queried the model for many more suggestions: on average 3.72 suggestions with highlighting and 2.91 without per email.
In contrast to the message-level suggestion condition, these suggestions were not used as often.
Participants only accepted 3.32 of them per email on average, and in 9 (8\%) interactions, no suggestions were used at all.
\rev{We did observe a difference between acceptance rates between the queries with and without highlighting: participants accepted 60.5\% of suggestions with highlighting and 36.7\% of suggestions without it.}
Finally, participants also contributed more tokens themselves in the final response compared to their counterparts who received \msg-level suggestions, as only 50.60\% of the tokens in the final replies originated from the suggestions they triggered (Figure~\ref{fig:eval-pct-human-all}).

\subsection{Writers' perceptions}
\label{sec:perceptions}

\begin{figure*}
    \centering
    \includegraphics[width=0.80\textwidth]{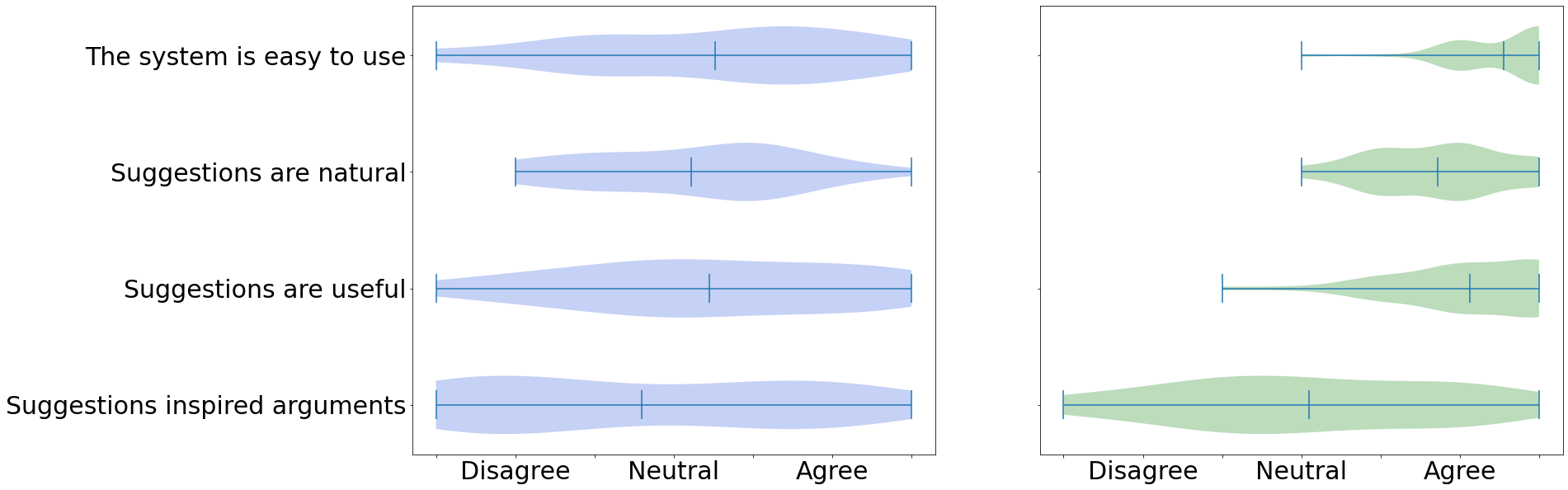}
    \caption{Participants who receive \msg-level suggestions (Right) tend to be more satisfied with the suggestions than participants in the sentence-level suggestions group (Left).}
    \label{fig:q1}
\end{figure*}

\begin{figure*}
    \centering
    \includegraphics[width=0.90\textwidth]{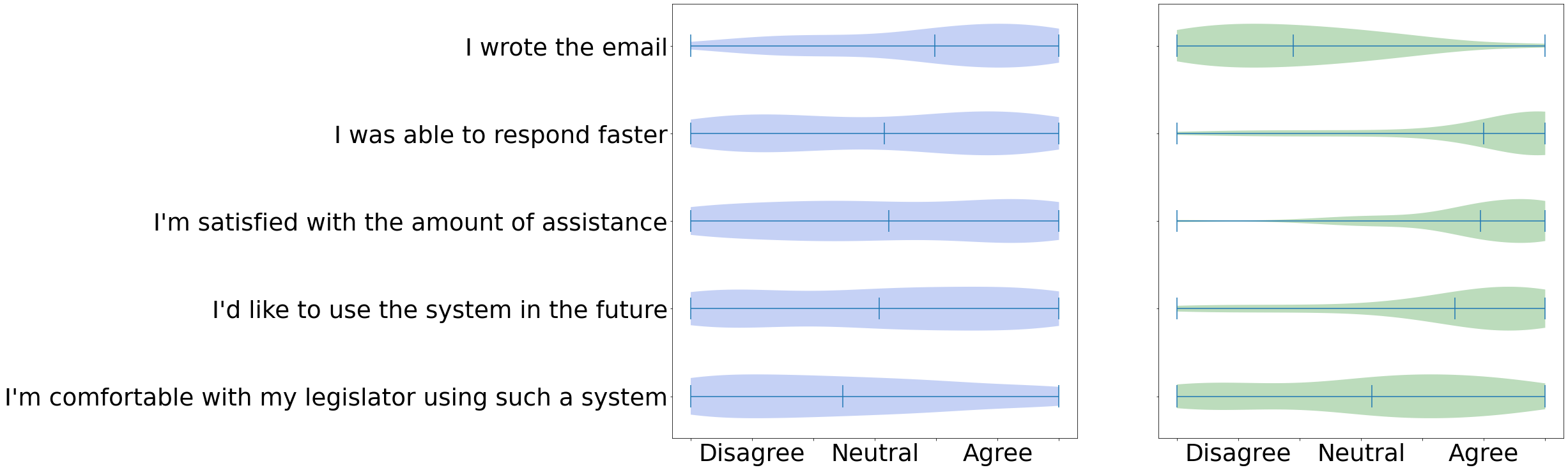}
    \caption{Participants who receive sentence-level suggestions (Left) retain a higher level of agency (Top row, ``I wrote the emails") than participants in the \msg-level suggestions group (Right).}
    \label{fig:q2}
\end{figure*}

To further understand how each type of assistance is perceived by the users, at the end of the experiment, we surveyed the participants about their perceptions of the suggestions they received and their level of comfort towards political communication mediated by the type of AI assistance they just experienced. 
The post-task survey consists of both Likert-scale questions and free-form feedback about their writing experience (See Appendix~\ref{appendix:survey} for the full list of questions).

\xhdr{Perceived helpfulness of the suggestions} Participants who received \msg-level suggestions generally agreed that the system is easy to use and that the suggestions they received were natural and useful (Figure~\ref{fig:q1}, Right).
However, participants in the sentence-level suggestions condition seemed to have diverging views, and did not rate the naturalness and usefulness of the suggestions as favorably (Figure~\ref{fig:q1}, Left).  

This contrast is also reflected in the free-form responses. %
Sentence-level suggestions are sometimes described as impersonal and not very natural:
{\it
\begin{itemize}
    \item[] ``It sounds a bit automated or kind of general sounding, but so do most politicians"
    \item[] ``Most of the suggestions came off as impersonally and artificially uber-patriotic."
\end{itemize}
}
%
%
However, participants who received \msg-level suggestions seem quite impressed with the naturalness of the suggestions they received:
{\it
\begin{itemize}
    \item[] ``It does sound like it were written by a human and is fully grammatically correct.  When it got it right, there were barely any modifications needed on my end. 
    \item[] ``I like how empathetic and personable the system is. At no point did I feel like these responses were from a machine. As such, I am curious to try the system out in my everyday life."
    \item[] ``It was quick and the suggested email was similar to what I would've written anyways."
\end{itemize}

}

%

A number of factors may have contributed to such differences. 
First, with the fixed token cap we use in our experiments, the generated suggestions may be cut short and not fully express an idea for the types of topics being discussed. 
Second, while the \msg-level suggestions have the full email as prompts, the sentence-level suggestions are generated with a more limited context and thus might be of lower quality. 
Future work may consider incorporating the \msg-level context, or even participants' interaction history, while offering response suggestions towards specific points to further improve the quality of suggestions.

We also notice that participants in both conditions felt rather neutral about the suggestions' capabilities in inspiring arguments they had not thought of (Figure~\ref{fig:q1}), pointing to an area for future improvement for the generation models.

\xhdrq{How did the suggestions help} Participants who received either type of assistance expressed how they liked that the suggestions served as starting points, as arguably the hardest part of the writing process is the beginning:

{\it 
\begin{itemize}
    \item[] ``I liked that it gave me suggestions for how to start out when I needed inspiration." 
    \item[] ``It was extremely handy especially when you dont know what to say or how to word your reply. "
    \item[] ``It is always easier to edit something than write it, even if the starting point is bad---these were solid though."
\end{itemize}
}


However, as text suggestions were presented in very different forms, participants made use of the suggestions in different ways. 
While participants who received sentence-level suggestions tended to find suggestions helpful in making them keep a more professional tone in their response:
{\it 
\begin{itemize}
    \item[] ``It was easier to keep a professional and political tone, and to quickly generate generic sentences."
    \item[] ``I like that it guided me to answer the letters in a professional manner."
\end{itemize}
}
%
%
Participants who receive \msg-level suggestions mainly commented on the usefulness of the draft serving as an outline for further editing:
{\it 
\begin{itemize}
    \item[] ``With a single click, I had an entire outline for an email, with minimal adjustments to be made."
    \item[] ``It gave a good base outline of how to respond that I could then use to expand upon and put emphasis on things that were really important to the topic."
\end{itemize}
}


\begin{figure*}
    \centering
    \includegraphics[width=0.95\textwidth]{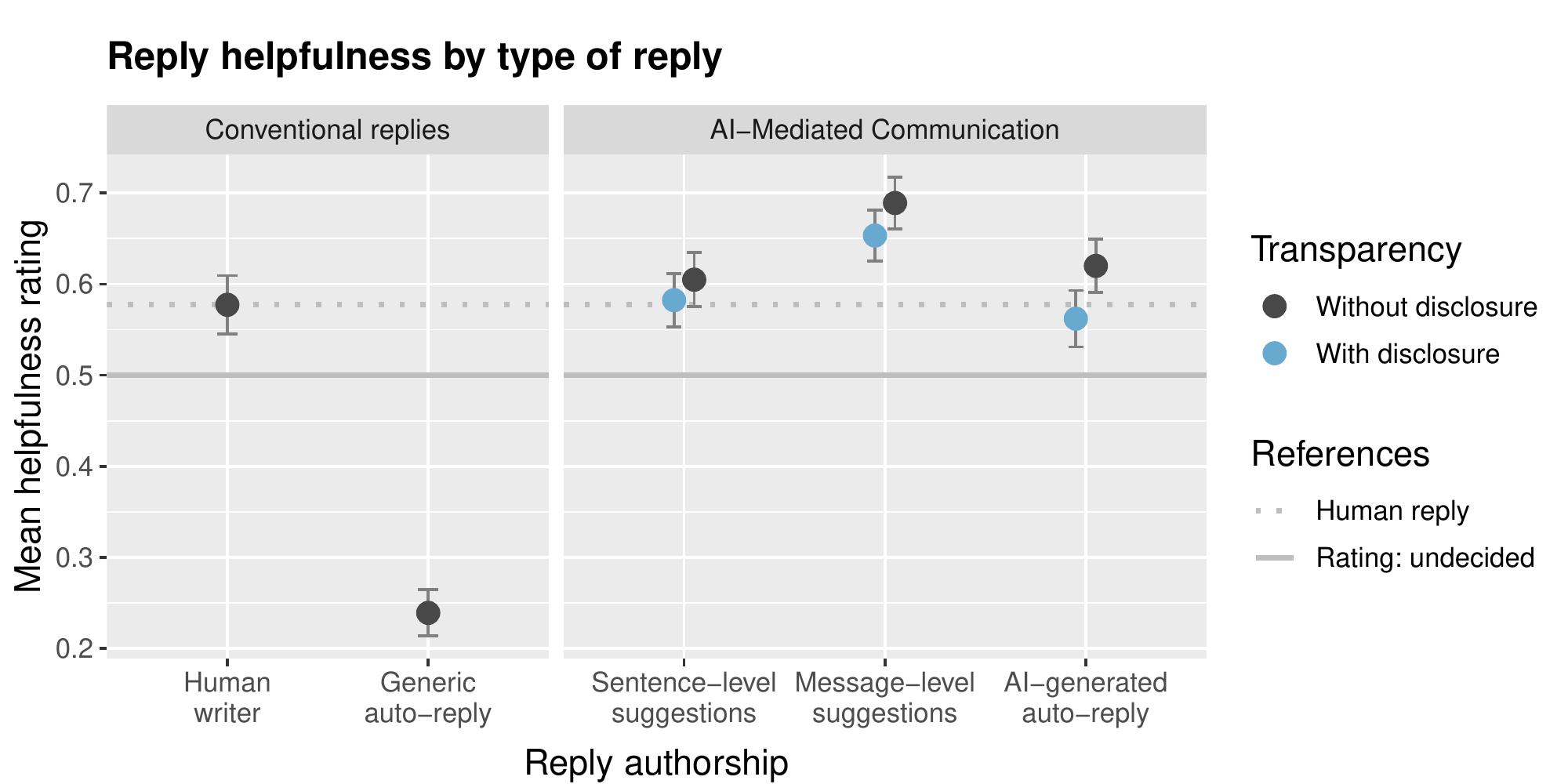}
    \caption{\textbf{Responses written with \msg-level suggestions were rated as significantly more helpful than those written with sentence-level suggestions and even than responses written without AI assistance.} N=500 judgements per data point. Error bars represent 95\% CIs. The Y-axis indicates the mean helpfulness ratings participants rated a reply with depending on the type of reply shown in the X-axis. Replies shown with an explicit disclosure of the AI involvement are color-coded in blue.}
    \label{fig:q1}
\end{figure*}

\xhdr{Writers' Agency} While participants in the sentence-level suggestions condition perceived that they retain substantial agency (Figure~\ref{fig:q2}, Left), 
participants who received \msg-level suggestions tended to think that they played a lesser role in drafting responses (Figure~\ref{fig:q2}, Right).  
This echoes our earlier observation that participants in the sentence-level condition contributed a much higher percentage of tokens than participants receiving \msg-level assistance. 
The autonomy granted to the AI has previously been identified as a key dimension in characterizing AI-mediated communication \cite{hancock_ai-mediated_2020}. The contrast in the perceived level of agency we observe in our experiment further demonstrates the need to clarify the desired level of agency for writers to retain in order to design appropriate assistance systems.  

\xhdr{Likelihood of future use} To explore potential receptions from both the writers' and the receivers' perspectives, we asked the participants not only how willing they would be to use such a system to respond to their emails, but also how comfortable they would be if their legislators were to use such a system to respond to their emails (Figure~\ref{fig:q2}). 
We find that participants tend to feel more hesitant about the platform if they were on the receiving end. The participants in the \msg-level suggestions condition expressed a rather strong willingness to use similar assistance to respond to their own emails (Figure~\ref{fig:q2}, Fourth row), but they did not feel as comfortable with their legislators using such a system to reply to them (Figure~\ref{fig:q2}, Last row). 
This suggests that beyond the effectiveness of the assistance, care must be taken for introducing and disclosing the use of such systems to people on both ends of the communication process to avoid tension and mistrust.

\subsection{Readers' perceptions}
\label{sec:helpfulness}


Following the main study, we conducted a follow-up study to understand how readers would perceive the replies written with the Dispatch system.
In the follow-up study, we took replies participants had written in the main study and asked a separate set of crowdworkers how helpful the replies were.
In addition to the replies written by the main experiment participants, we also evaluated the helpfuless of message-level suggestions generated by GPT-3 as described in the main experiment without any human editing.
To this set of replies, we added a sample of generic auto-replies that legislators sent to real-world inquiries in a previous field study from the Cornell Tech Policy Institute.

We recruited 1,000 participants on Prolific \cite{palan2018prolific} to evaluate these replies to legislative inquiries.
We developed a mock-up of an email conversation displaying both the citizen concerns that main experiment participants had responded to as well as a specific reply.
Each participant rated one reply written with sentence-level suggestions, one reply written with message-level suggestions, one reply generated by GPT-3 without human editing as well as one reply that was either a generic auto-reply or written by a human without AI assistance.
In addition, half of the participants saw a disclosure label stating that  ``Elements of this reply were generated by an AI communication tool.'' when they saw replies that had been written either by GPT-3 itself or with the help of GPT-3.
For each reply, we asked participants whether they agreed with the statement that ``The reply is helpful and reasonable'' on a 5-point Likert-scale from ``Disagree'' to ``Agree''. For a statistical analysis, we converted their responses to a numeric scale from 0 to 1 respectively and conducted a linear regression analysis with human-written replies as the baseline. 

The results are shown in Figure \ref{fig:q1}. 
When evaluating the replies participants had written in the main study, participants in the follow-up task indicated that replies written with message-level suggestions (M=0.69, shown central in the right panel) were more helpful than those replies people had written without AI assistance
(left in left panel, $M=0.57$, $t_{(973)}=-5.11$, $p<0.0001$). Replies that were written with sentence-level suggestions (M=0.60, left in right panel) were seen as less helpful than those written with message-level suggestions and similarly helpful to those written without AI assistance. Replies that GPT-3 generated without human supervision (right in right panel) were seen as slightly more helpful than replies that people had written without AI assistance ($M=0.62$, $t_{(979)}=-1.92$, $p=0.054$). In comparison, the generic auto-replies (right in left panel) that busy legislators may send to cope with an overwhelming volume of inquiries were rated as very unhelpful ($M=0.24$, $t_{(942)}=16.2$, $p<0.0001$). Explicitly disclosing the involvement of AI in the reply generation (shown in blue) may have reduced the perceived helpfulness of replies generated with message-level suggestions ($M=0.65$, $t_{(998)}=1.75$, $p=0.08$) and of replies autonomously generated by GPT-3 ($M=0.56$, $t_{(995)}=2.67$, $p=0.07$). However, even when the AI involvement was explicitly disclosed, replies written with message-level suggestions were seen as significantly more helpful than replies written with sentence-level suggestions.

\subsection{Characteristics of the responses}
\label{sec:responses}

While we have discussed the effects of suggestion type on the writing process, we are also interested in how the suggestion type affects the final written product itself.
In particular, we investigate three aspects:

\xhdr{Length} We compare the number of words in the responses under different assistance conditions. We find that participants in the control condition produced the longest responses, averaging 115.8 words.  
The average length of responses from the sentence-level suggestions condition and \msg-level suggestions condition were both significantly shorter than responses from the control condition, at an average of 90.5 words ($t_{(238)}=-4.94$, $p < 0.001$) and 88.0 words ($t_{(238)}=-4.29$, $p < 0.001$) respectively.\footnote{We set {\it max\_tokens} to 200 for generating \msg-level suggestions, but the generated suggestions have, on average, 74.3 words.} 
This is counter-intuitive as one might expect participants with access to suggestions to write more, but this is not the case.
The reasons for this result are unclear, but one possibility is that participants in the \msg-level suggestions condition anchored very strongly to the length of the generated suggestions, and were less likely to add more content. 
As for participants in the sentence-level suggestions condition, they might have expended additional energy in deciding where to trigger suggestions and choosing which suggestions to use, leading them to spend less time writing.\footnote{It is also possible that the suggestions in both settings packed more information into a smaller number of words while the human writers were unnecessarily verbose.} 

\xhdr{Grammaticality} We compared the grammaticality of responses written under different conditions. To do this, we computed the error rate of the responses as the number of grammatical errors divided by the number of words in the response. 
Following the methods from prior work \cite{Lee2022coauthor}, we used LanguageTool to identify grammatical errors in the responses.\footnote{We use the Python wrapper for computation: \url{https://github.com/jxmorris12/language_tool_python}.} 
Similar to previous studies \cite{dou_is_2022, Lee2022coauthor}, we find that responses from the message-level suggestions condition have the lowest error rate, averaging 0.158 errors per word. The responses from the sentence-level condition have a slightly higher average error rate, at 0.165 errors per word. The purely human-written control responses ended up having the highest error rate of 0.176 errors per word, \rev{which is significantly higher to both the sentence-level condition ($t_{(238)}=2.64$, $p < 0.01$) and the \msg-level condition ($t_{(238)}=4.31$, $p < 0.001$).} 

\xhdr{Vocabulary diversity} Vocabularity diversity is a proxy for how engaging or interesting the generations are, and to measure it, we use the distinct-2 score \cite{li_diversity-promoting_2016}, i.e., the number of unique bigrams / the total number of words in the response.
NLP model generations tend to be less diverse than human written text \cite{li_diversity-promoting_2016, Welleck2020NeuralTG}, which is reflected in our results: the control responses have higher distinct-2 scores ($M = 0.944$) than both the responses from sentence-level suggestions condition ($M = 0.928$) and those from message-level suggestions condition ($M = 0.934$), although only the difference between the control responses and the responses from the sentence-level suggestions condtition is significant ($t_{(238)} = 2.91$, $p < 0.01$). 

%


\section{Design Implications}
\label{sec:implications}


In this work, we explore the effects of sentence-level vs. \msg-level suggestions in assisting users in email communication. 
We observe that different forms of suggestions lead to substantially different writing processes: participants receiving \msg-level suggestions mostly edited the drafts presented to them, skipping the first two steps in the traditional ``outline-draft-edit" process \cite{griffith_english_1977} that participants who receive sentence-level suggestions still seem to go through.  
As a result, participants who received \msg-level suggestions finished their responses faster while participants who received sentence-level suggestions retained a higher sense of agency in the process. 

This contrast, together with other observations we notice in our experiment, suggests that as more technically advanced options become feasible, each with its own advantages and shortcomings, it is all the more important to better understand the needs and specifications of the particular communication circumstance to design task-appropriate assistance systems. 
Below, we outline a few technical options that may be considered and adjusted according to the communication circumstance. 

\xhdr{Unit of suggestion} Suggestions can be offered at different units and lengths. While we experiment with suggestions at the level of short sentences and messages, there are intermediate forms as well, e.g., longer sentences or even paragraphs. 
As we have observed, this choice affects not just the text entry efficiency, but more fundamentally, the writing process itself.\footnote{In fact, prior work suggests that even the difference between word-level and phrase-level suggestions may have such effect \cite{Arnold2016phrases_vs_words}.}
The more complete a draft the suggested text offers, the more the users' focus may be shifted towards editing and less towards outlining and drafting, implying different degrees of delegation of the writing process. 
%
%
Hence, finding the appropriate suggestion unit requires finding the sweet spot that tailors both to the communication topic, as different topics may take different amount of text to fully develop and express an idea, and also to communicators' willingness to delegate the task. For instance, prior research reports that people prefer less machine assistance when writing a birthday card to their mothers compared to when responding to mundane work emails \cite{lubars_ask_2019}.

\xhdr{Availability of options} Participants who received sentence-level suggestions were provided with five candidate options whenever they triggered a suggestion.
The availability of choices could potentially allow users more flexibility and increase the chance of offering at least one useful suggestion. 
In fact, some of the participants receiving \msg-level suggestions expressed their interest in receiving more candidate responses, e.g., {\it ``maybe have it give you a choice of 2 or 3 different responses"}). 
However, reading and deciding between candidate suggestions can sometimes be distracting and can take a considerable amount of time, as we have seen earlier (i.e. in Figure~\ref{fig:completion-time}).
%
Furthermore, offering choices is perhaps only beneficial when a set of diverse and complementary suggestions can be generated.
In our experiment, the sentence-level suggestions were sometimes too similar, leading to complaints such as {\it ``some of the suggestions were repetitive, or out of context"} and {\it ``The responses were too similar. Most began with `I agree!'"}. 
In addition, in communication settings when we expect a relatively narrow range of possible responses, like answering a factual question, having multiple options may not be needed or wanted.

\xhdr{Generation model} \rev{In this work, we used the best available model at the time (GPT-3) to generate both sentence-level and \msg-level suggestions, as we are primarily concerned with how humans interact with different types of assistance and we hope to make as fair a comparison as possible.
Some of the challenges we observe with generating sentence-level suggestions, i.e., limited input context and limited space to fully develop an argument, are inherent to the task itself and would likely remain even if a different model were used.}

\rev{Observations and limitations from our studies also point to a number of ways models could be improved to better faciliate such communication process. For instance, the priming effects we observe---participants who recieve \msg-level assistance write shorter messages than those in the control group because the suggestions are shorter---suggest that fine-tuning generation models to exhibit specific properties, e.g. a particular length range or a formal tone could be helpful. 
Furthermore, while we have made the distinction based on party affiliation when generating suggestions, legislators can have much more finer-grained differences in their policy stances and communication styles. Personalizing suggestions based on policy stances and communication styles of the legislators is another important avenue for future work.}

\xhdr{Beyond text suggestions} While we explore assistance options involving text suggestions, communication assistance systems can help in more aspects of the writing process and offer more than merely generating suggestions for content.
Quoting from our participants' suggestions, additional assistance could range from {\it ``highlighting key points and adding blank spaces to share personal opinions and ideas''} to help with outlining, {\it ``making it easier to see which text the system created and which text was typed by me"} to help with reviewing, tracking {\it ``responses I made earlier about a similar topic"} for future use, or providing related contextual information by generating {\it ``a quick tutorial on the subject"}.

\xhdr{Disclosure of assistance} While we have focused on the writers' perspective, it is important to remember that successful communication is not just about replying to all of the emails.
In many cases, such as in legislator-constituent communication, it is more important to build trust and understanding between people who are communicating.
In our experiment, even participants who expressed relatively strong interest in using assistance systems to reply to emails were hesitant about having their legislators use the same system. 
As such, if such a system were to be incorporated into staffers' workflow, it is important to consider how to disclose and explain its use to avoid further friction and mistrust between constituents and legislators.

These decisions do not have purely technical solutions. While we attempt to lay out feasible technical options, ultimately, we hope to facilitate the communication processes, and it should be up to the communicators themselves---i.e., staffers, legislators, and constituents in the context of legislator-constituent communication---to make the important value judgments on what they feel comfortable delegating to assistance systems.

\section{Conclusion}
\label{sec:conclusion}

In this work, we explored two assistance options enabled by the capability of recent large language models to generate long, natural-sound suggestions: sentence-level suggestions and \msg-level suggestions.  
To understand the trade-offs between these two types of suggestions, we conducted an online experiment via \dispatch, a platform we built to simulate the scenario of staffers from legislative offices responding to constituents' concerns. 
The results show that different forms of suggestions can affect the participants' writing experience in multiple dimensions. 
For instance, participants receiving \msg-level suggestions mainly edited the suggested responses and were able to complete the task significantly faster, 
while participants receiving sentence-level suggestions retained a higher sense of agency and contributed more original content. 
%
We discussed the implications of our observations for designing assistance systems tailored specifically to the communication circumstance.   
%

Different communication circumstances have different objectives and demands.
Efficiency may be at the core of customer services communications whereas developing trust and credibility is critical for legislator-constituent communication. 
\rev{This work has provided an initial proof-of-concept that we hope will encourage further exploration of communication assistance systems beyond the specific domain studied here. 
For example, while we targeted users fluent in English, these systems could be even more beneficial to those less proficient in the language.
Studying the utility and effectiveness for non-native English speakers would be a fruitful extension of this research.
Recent work has demonstrated the capacity of language models to parse ideological nuance, which would be especially important in environments with political polarization where espousing the ``wrong'' position could alienate voters, although that dynamic was outside the scope of the current study and should be considered for follow-on research.}

\rev{To conclude, in this work we shed light on the factors relevant for writing assistance systems for legislator-constituent communication. We hope our work encourages further studies towards designing task-appropriate communication assistance systems.}


\xhdr{Acknolwedgement} We thank the anonymous reviewers for their helpful comments and Nikhil Bhatt, Gloria Cai, Paul Lushenko, Meredith Moran, Tanvi Namjoshi, Shyam Raman, Aryan Valluri, Ella White for their help with internal testing.  This research was supported by a New Frontier Grant from the College of Arts and Sciences at Cornell.





\bibliographystyle{ACM-Reference-Format}
\bibliography{tech-policy, tech-hci}

\appendix

\section{Task details}

\subsection{Instructions}
\label{appendix:instructions}

Instructions for how different types of suggestions can be triggered are shown below. 

{\it Sentence-level suggestions} 
\begin{quote}
    When drafting your responses, you can trigger two types of response suggestions:
    
    1. {\bf HIGHLIGHT} a sentence in the letter and {\bf TYPE "@"} in the editor to trigger suggestions that directly respond to the sentence.  
    
    2. {\bf TYPE "@"} in the editor without highlighting to trigger suggestions for how to continue what you're writing. 
\end{quote}

\rev{{\it Message-level suggestions}}
\begin{quote}
    To trigger a suggested reply from the AI assistant, press the {\bf Generate} button under the left panel. You can then edit the generated email to your liking. 
\end{quote}

\subsection{Survey questions}
\label{appendix:survey}

1. Choose the degree to which you agree with the following statements: 
\begin{itemize}
    \item The system was easy to use. 
    \item The system's suggestions sound natural.  
    \item The system's suggestions were useful. 
    \item The system's suggestions inspired me to include points I hadn't thought of. 
\end{itemize}

2. Choose the extent to which you agree with the following statements:
\begin{itemize}
    \item I wrote the emails. 
    \item I was able to respond to emails faster than normal. 
    \item I'm satisfied with the amount of assistance I received from the system. 
    \item I would like to respond to emails using this system in the future. 
    \item I would be comfortable with my legislator using a system like this to respond to my emails. 
\end{itemize}

3. What did you like about the experience of responding to emails using the system? 

4. What would you change about the system to improve your experience responding to emails? 

\section{Additional Results}


%
\xhdr{Adaptation} In both of these conditions, we were also interested in how the participants' use of the system changed as they drafted each message.
Table \ref{tab:pct_human} shows the percentage of tokens from the suggestion or from the participant over the first, second, and third replies written.
In the \rev{message-level} suggestions condition, participants included slightly more suggestion tokens in the first message than the later ones, while in the sentence-level suggestions condition, suggestions were used more in the second message.
That said, these differences are small and suggest that participant behavior was consistent across all interactions.
Future work might investigate each participant drafting more messages to see if there is any adaptation behavior.

\begin{table}[]
    \centering
    \begin{tabular}{lrrr}
    \toprule
    & \multicolumn{3}{c}{Letter Index}\\
    Condition &     \multicolumn{1}{c}{1} &     \multicolumn{1}{c}{2} &     \multicolumn{1}{c}{3} \\
    \midrule
    control  &  $100.00\pm0.00$ &  $100.00\pm0.00$ &  $100.00\pm0.00$ \\
    sentence &  $51.28\pm28.21$ &  $48.91\pm31.06$ &  $51.10\pm30.19$ \\
    email    &  $20.47\pm20.86$ &  $24.25\pm24.54$ &  $24.51\pm23.36$ \\
    \bottomrule
    \end{tabular}
    \caption{The percentage of tokens in the reply that come from the human participant across the three emails each participant wrote.}
    \label{tab:pct_human}
\end{table}

\begin{table}[]
    \centering
    \begin{tabular}{lrrr}
    \toprule
    email id &  control &  message-level &  sentence-level \\
    \midrule
    0                  &   100.00 &  20.66 &     41.31 \\
    1                  &   100.00 &  26.81 &     43.96 \\
    2                  &   100.00 &   7.77 &     49.70 \\
    3                  &   100.00 &  31.90 &     66.07 \\
    4                  &   100.00 &  11.82 &     38.36 \\
    5                  &   100.00 &  29.71 &     40.40 \\
    6                  &   100.00 &  17.30 &     45.58 \\
    7                  &   100.00 &  35.92 &     66.69 \\
    8                  &   100.00 &  21.63 &     49.27 \\
    9                  &   100.00 &  26.99 &     35.96 \\
    10                 &   100.00 &  20.73 &     54.91 \\
    11                 &   100.00 &  25.69 &     72.92 \\
    \midrule
    \textbf{all}                &   100.00 &  24.25 &     49.40 \\
    \bottomrule
    \end{tabular}
    \caption{Average percentage of human-written text for each message across all the conditions. All of the control replies are human-written while almost a quarter of the message-level replies and half of all sentence-level replies are human-written.}
    \label{tab:by_email}
\end{table}

\begin{figure}
    \centering
    \includegraphics[width=0.4\textwidth]{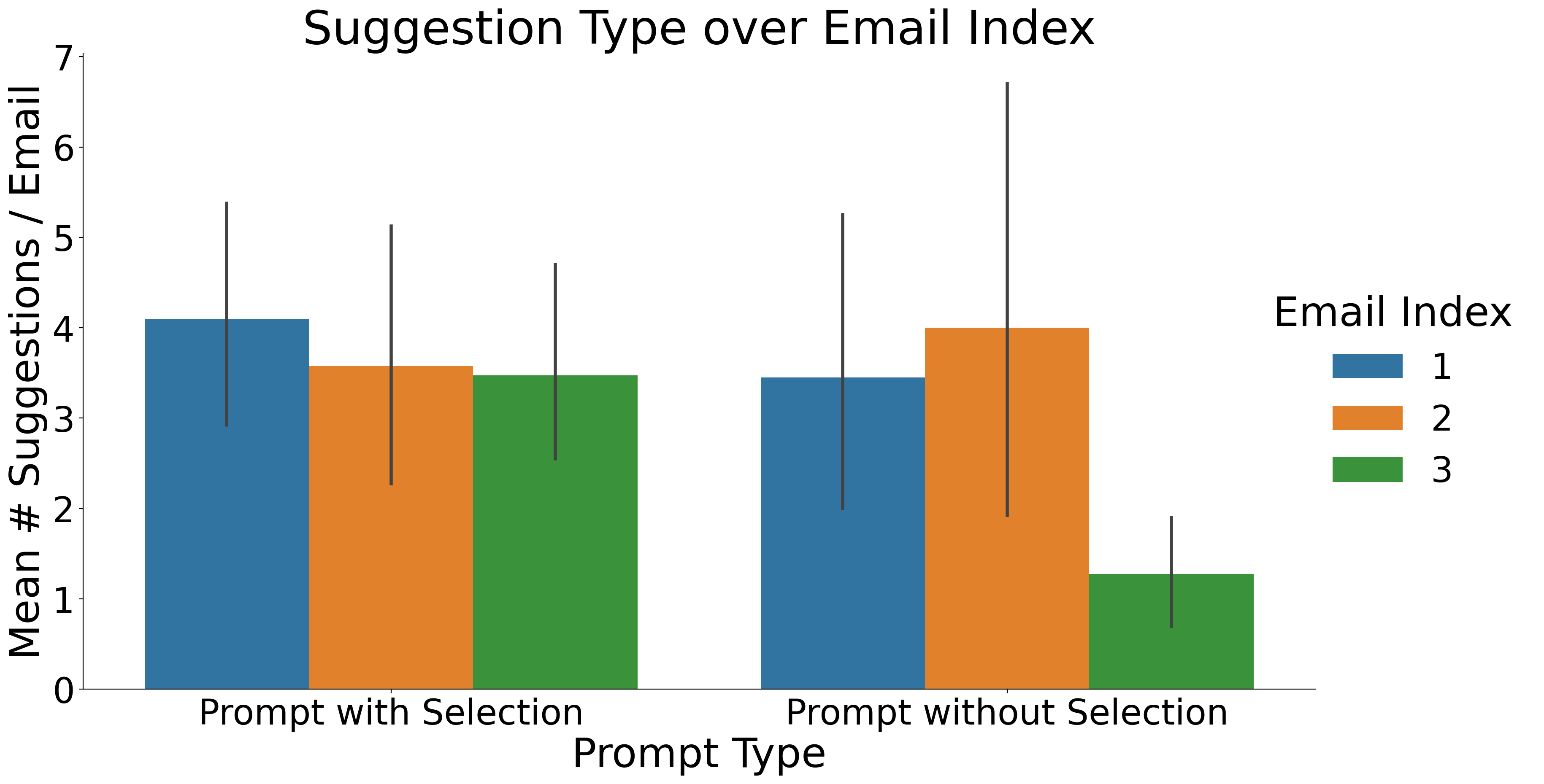}
    \caption{The average number of suggestions prompted with and without highlighting a portion of the message across the three emails each participant wrote.}
    \label{fig:sentence_suggest_types}
\end{figure}

However, there was a difference in the types of suggestions triggered in the sentence-level condition.
When writing the final message, participants more often prompted the model for suggestions without highlighting any text (Figure \ref{fig:sentence_suggest_types}).
%


\xhdr{Effect of message}
One concern we might have is that the different messages loaned themselves to better suggestions.
To investigate this, we looked at the percentage of human-written tokens for each of the messages across all of the conditions (Table \ref{tab:by_email}).
We found that overall, there was not too much variation among the ten samples of each message, with messages 3, 7, and 11 (for the sentence-level suggestions) having the most human-written tokens across both conditions.

\end{document}